\documentclass[journal]{IEEEtran}
\usepackage{cite}
\usepackage{times}
\usepackage{epsfig}
\usepackage{graphicx}
\usepackage{amsmath}
\usepackage{amssymb}
\usepackage{booktabs}
\usepackage{color}
\usepackage{soul}
\usepackage{multirow}
\usepackage{url}
\ifCLASSINFOpdf
\else
\fi
\begin{document}
%
% paper title
% Titles are generally capitalized except for words such as a, an, and, as,
% at, but, by, for, in, nor, of, on, or, the, to and up, which are usually
% not capitalized unless they are the first or last word of the title.
% Linebreaks \\ can be used within to get better formatting as desired.
% Do not put math or special symbols in the title.
\title{Learning Guided Convolutional Network for \\ Depth Completion}
%
%
% author names and IEEE memberships
% note positions of commas and nonbreaking spaces ( ~ ) LaTeX will not break
% a structure at a ~ so this keeps an author's name from being broken across
% two lines.
% use \thanks{} to gain access to the first footnote area
% a separate \thanks must be used for each paragraph as LaTeX2e's \thanks
% was not built to handle multiple paragraphs
%

\author{Jie~Tang,~\IEEEmembership{Student Member,~IEEE,}
        Fei-Peng~Tian,~\IEEEmembership{Student Member,~IEEE,}
        Wei~Feng,~\IEEEmembership{Member,~IEEE,}
        Jian~Li,~\IEEEmembership{Member,~IEEE,}
        and~Ping~Tan,~\IEEEmembership{Member,~IEEE}% <-this % stops a space
\thanks{J. Tang and F.-P. Tian are joint first authors contributing equally to this work. 
J. Li is the corresponding author. Email: lijian@nudt.edu.cn.}
\thanks{This work is done when J. Tang and F.-P. Tian are visiting at GrUVi Lab, School of Computing Science, Simon Fraser University, Burnaby, BC, Canada.}

\thanks{J. Tang and J. Li are with the College of Intelligence Science and Technology,
National University of Defense
Technology, Changsha 410073, China.}
\thanks{F.-P. Tian and W. Feng are with the School of Computer
Science and Technology, College of Intelligence and Computing, Tianjin University, TianJin 300350, China, and the Key Research Center for Surface Monitoring and Analysis of Cultural Relics (SMARC), State Administration of Cultural Heritage, Beijing, China.}% <-this % stops a space
\thanks{P. Tan is with the School of Computing Science, Simon Fraser University, Burnaby, BC, Canada.}
}

% note the % following the last \IEEEmembership and also \thanks - 
% these prevent an unwanted space from occurring between the last author name
% and the end of the author line. i.e., if you had this:
% 
% \author{....lastname \thanks{...} \thanks{...} }
%                     ^------------^------------^----Do not want these spaces!
%
% a space would be appended to the last name and could cause every name on that
% line to be shifted left slightly. This is one of those "LaTeX things". For
% instance, "\textbf{A} \textbf{B}" will typeset as "A B" not "AB". To get
% "AB" then you have to do: "\textbf{A}\textbf{B}"
% \thanks is no different in this regard, so shield the last } of each \thanks
% that ends a line with a % and do not let a space in before the next \thanks.
% Spaces after \IEEEmembership other than the last one are OK (and needed) as
% you are supposed to have spaces between the names. For what it is worth,
% this is a minor point as most people would not even notice if the said evil
% space somehow managed to creep in.

% The paper headers
% \markboth{Journal of \LaTeX\ Class Files,~Vol.~14, No.~8, August~2015}%
% {Shell \MakeLowercase{\textit{et al.}}: Bare Demo of IEEEtran.cls for IEEE Journals}
\markboth{SUBMISSION TO IEEE TRANSACTIONS ON IMAGE PROCESSING, 2019}%
{Shell \MakeLowercase{\textit{et al.}}: Bare Demo of IEEEtran.cls for IEEE Journals}

% The only time the second header will appear is for the odd numbered pages
% after the title page when using the twoside option.
% 
% *** Note that you probably will NOT want to include the author's ***
% *** name in the headers of peer review papers.                   ***
% You can use \ifCLASSOPTIONpeerreview for conditional compilation here if
% you desire.

% If you want to put a publisher's ID mark on the page you can do it like
% this:
%\IEEEpubid{0000--0000/00\$00.00~\copyright~2015 IEEE}
% Remember, if you use this you must call \IEEEpubidadjcol in the second
% column for its text to clear the IEEEpubid mark.

% use for special paper notices
%\IEEEspecialpapernotice{(Invited Paper)}

% make the title area
\maketitle

% As a general rule, do not put math, special symbols or citations
% in the abstract or keywords.
% \begin{abstract}
% The abstract goes here.
% \end{abstract}
%!TEX root = root.tex

%%%%%%%%% ABSTRACT
\begin{abstract}
    Dense depth perception is critical for autonomous driving and other robotics applications.
    However, modern LiDAR sensors only provide sparse depth measurement.
    It is thus necessary to complete the sparse LiDAR data, where a synchronized guidance RGB image is often used to facilitate this completion. 
    Many neural networks have been designed for this task.
    However, they often na\"{\i}vely fuse the LiDAR data and RGB image information by performing feature concatenation or element-wise addition. 
    Inspired by the guided image filtering, we design a novel guided network to predict kernel weights from the guidance image.
    These predicted kernels are then applied to extract the depth image features.
    In this way, our network generates \emph{content-dependent} and \emph{spatially-variant} kernels for multi-modal feature fusion.
    Dynamically generated spatially-variant kernels could lead to prohibitive GPU memory consumption and computation overhead.
    We further design a convolution factorization to reduce computation and memory consumption.
    The GPU memory reduction makes it possible for feature fusion to work in multi-stage scheme.
    We conduct comprehensive experiments to verify our method on real-world outdoor, indoor and synthetic datasets.
    Our method produces strong results.
    It outperforms state-of-the-art methods on the NYUv2 dataset and ranks 1st on the KITTI depth completion benchmark at the time of submission.
    It also presents strong generalization capability under different 3D point densities, various lighting and weather conditions as well as cross-dataset evaluations.
    The code will be released for reproduction.
\end{abstract}

% Note that keywords are not normally used for peerreview papers.
\begin{IEEEkeywords}
Depth completion, depth estimation, guided filtering, multi-modal fusion, convolutional neural networks.
\end{IEEEkeywords}

% For peer review papers, you can put extra information on the cover
% page as needed:
% \ifCLASSOPTIONpeerreview
% \begin{center} \bfseries EDICS Category: 3-BBND \end{center}
% \fi
%
% For peerreview papers, this IEEEtran command inserts a page break and
% creates the second title. It will be ignored for other modes.
\IEEEpeerreviewmaketitle

%!TEX root = root.tex

\section{Introduction}
\IEEEPARstart{D}{ense} depth perception is critical for many robotics applications, such as autonomous driving or other mobile robots.
Accurate dense depth perception of the observed image is the prerequisite for solving the following tasks such as obstacle avoidance, object detection or recognition
and 3D scene reconstruction.
While depth cameras can be easily adopted in indoor scenes, outdoor dense depth perception mainly relies on stereo vision or LiDAR sensors.
Stereo vision algorithms~\cite{sgm,middleburry,cnn_stereo, efficient_stereo} still have many difficulties in reconstructing thin and discontinuous objects.
So far, LiDAR sensors provide the most reliable and most accurate depth sensing and have been widely integrated into many robots and autonomous vehicles.
However, current LiDAR sensors only obtain sparse depth measurements, e.g. 64 scan lines in the vertical direction.
Such a sparse depth sensing is insufficient for real applications like robotic navigation.
Thus, estimating dense depth map from the sparse LiDAR input is of great value for both academic research and industrial applications.

Many recent works~\cite{sparsity_cnn, deep_lidar, self_supervised}  on this topic take deep learning as approach and exploit an additional synchronized RGB image for depth completion.
These methods have achieved significantly improvements over conventional approaches~\cite{dense_disparity,sparse_sample,defense_classical}.
For example,  Qiu et al.~\cite{deep_lidar} train a network to estimate surface normal from both the RGB image and LiDAR data and further use the recovered surface normal to guide depth completion.
Ma et al.~\cite{self_supervised} exploit photo-consistency between neighboring video frames for depth completion.
Jaritz et at.~\cite{depth_comp_semantic} adopt a depth loss as well as a semantic loss for supervision.
Despite the different methods proposed by these works, they basically share the same scheme in multi-modal feature fusion.
Specifically,
these works adopt the operation like concatenation or element-wise addition to fuse the feature vectors from sparse depth and RGB image together directly for further processing.
However, the commonly used concatenation or element-wise addition operation is not such appropriate when considering the heterogenous data and the complex environments.
The potentiality of RGB image as guidance is difficult to be fully exploited by applying in such a simple way.
In contrast, we suggest a more sophisticated fusion module to improve the performance of the depth completion task.

\begin{figure*}
\begin{center}
\includegraphics[width=0.98\textwidth]{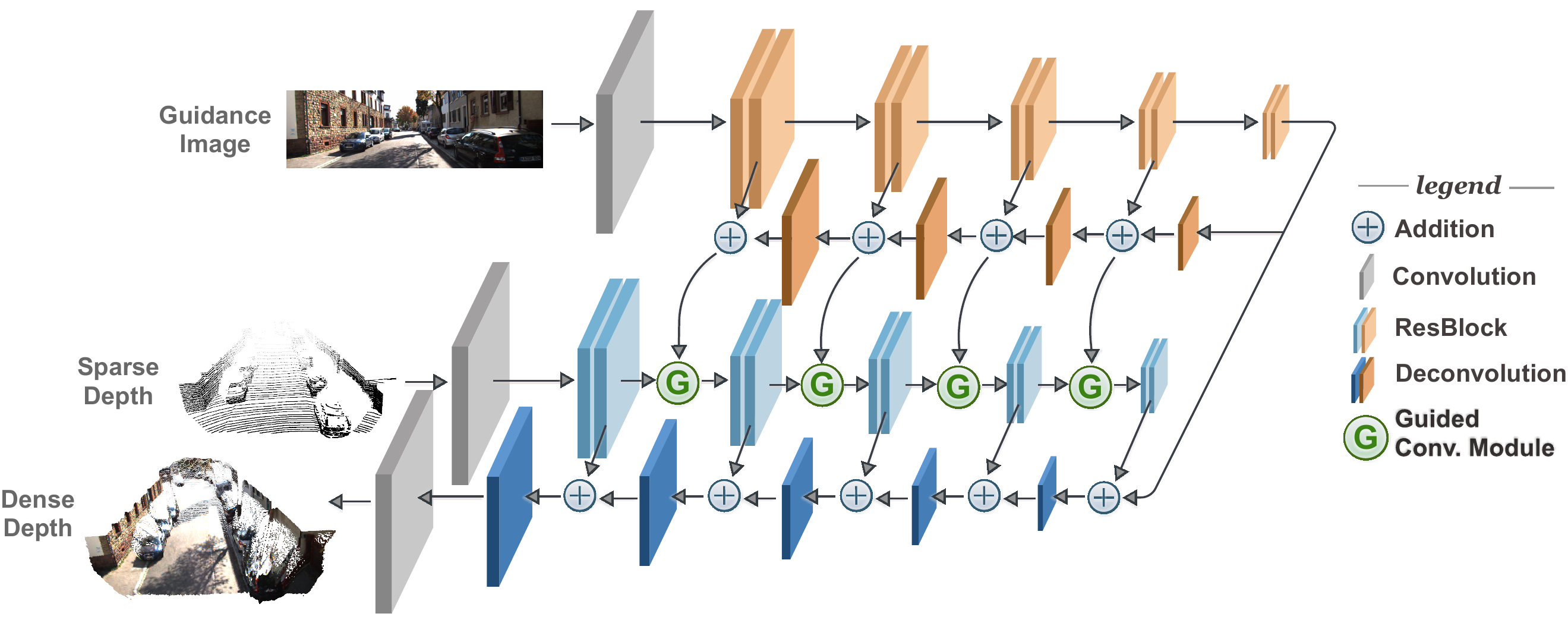}
\end{center}
   \caption{{\bf The proposed network architecture.}
   The whole network architecture includes two sub-networks: GuideNet in orange and DepthNet in blue.
   We add a standard convolution layer at the beginning of both GuideNet and DepthNet as well as the end of DepthNet.
   The light orange and blue are the encoder stages, while corresponding dark ones are decoder stage of GuideNet and DepthNet, respectively.
   The ResBlock represents the basic residual block structure with two sequential $3\times3$ convolutional layers from~\cite{res_net}. }
\label{fig:net_arch}
\end{figure*}

Our work is inspired by the guided image filtering~\cite{guided_filter, bilateral_filter}.
In guided image filtering, the output at a pixel is a weighted average of nearby pixels, where the weights are functions of the guidance image.
This strategy has been adopted for generic completion/super-resolution of RGB and range images~\cite{joint_upsampling, spatial_depth_range, joint_geodesic_upsampling}.
Inspired by the success of guided image filtering, we seek to learn a guided network
to automatically generate \emph{spatially-variant} convolution kernels according to the input image and then apply them to
extract features from sparse depth image by our guided convolution module.
Compared with the hand-crafted function for kernel generation in guided image filtering~\cite{guided_filter},
our end-to-end learned network structure has a potential to produce more powerful kernels with agreement of scene context for depth completion. 
Compared with standard convolutional module, where the kernel is \emph{spatially-invariant} and pixels at all the positions share the same kernels,
our guided convolutional module has spatially-variant kernels that are automatically generated according to the content.
Thus, our network is more powerful to handle various challenging situations in depth completion task.

An obvious drawback of using spatially-variant kernels is the large GPU memory consumption, which is also the original motivation of parameter sharing in the convolutional neural network.
Especially when applying the spatially-variant convolution module in the multi-stage fusion for depth completion,
the massive GPU memory consumption is even unaffordable for computational platforms
(See subsection~\ref{subsec:conv_fact} for memory and computation discussion).
Thus, it's non-trivial to look for a practical way to make the network available.
Inspired by recent network compression technique~\cite{MobileNets}, we factorize the convolution operation in our guided convolution module to two stages,
a \emph{spatially-variant} channel-wise convolution stage and a \emph{spatially-invariant} cross-channel convolution stage.
By using such a novel factorization,
we get an enormous reduction of GPU memories such that the guided convolution module can be integrated with the powerful encoder-decoder network in multi-stages in a modern GPU device.

The proposed method is evaluated on both outdoor and indoor datasets, from real-world and synthetic scenes.
It outperforms the state-of-the-art methods on KITTI depth completion benchmark and rank 1st at the time of paper submission.
Comprehensive ablation studies demonstrate the effectiveness of each component and the fusion strategy used in our method.  
Compared with other depth completion methods, our method also achieves the best performance on the indoor NYUv2 datset. 
Last but not least, our model presents strong generalization capability under different depth point densities, various lighting and weather conditions as well as cross-dataset~evaluations.
Our code will be released at \url{https://github.com/kakaxi314/GuideNet}.

%!TEX root = root.tex

\section{Related Work}
Depending on whether there is an RGB image to guide the depth completion, previous methods can be roughly divided into two categories: depth-only methods and image-guided methods.
We briefly review these techniques and other literatures relevant to our network design.

{\bf Depth-only Methods} These methods use a sparse or low-resolution depth image as input to generate a full-resolution depth map.
Some early methods reconstruct dense disparity maps~\cite{dense_disparity} or depth maps~\cite{sparse_sample} based on the compressive sensing theory~\cite{dense_disparity} or a combined wavelet-contourlet dictionary~\cite{sparse_sample}.
Ku et al.~\cite{defense_classical} use a series of hand-crafted conventional operators like dilation, hole closure, hole filling, and blurring, etc., to transform sparse depth maps into dense.
More recently, deep learning based approaches demonstrate promising results.
Uhrig et al.~\cite{sparsity_cnn} propose a sparsity invariant CNN to deal with sparse data or features by using an observation mask.
Eldesokey et al.~\cite{propagating_confidences} solve depth completion via generating a full depth as well as a confidence map with normalized convolution.
Chodosh et al.~\cite{compress_sensing_convolution} combine compressive sensing with deep learning for depth prediction.
The main focus of these methods is to design appropriate operators,
e.g. sparsity invariant CNN~\cite{sparsity_cnn}, to deal with sparse inputs and propagate these spare information to the whole image.

In terms of depth super-resolution, some methods exploit a database~\cite{patch_super_res} of paired low-resolution and high-resolution depth image patches or
self-similarity searching~\cite{depth_rigid_similarity} to generate a high resolution depth image.
Some methods~\cite{var_super_edge, joint_super_denoise} further propose to solve depth super-resolution  by dictionary learning.
Riegler et al.~\cite{atgv_net} use a deep network to produce a high-resolution depth map as well as depth discontinuities and feed them into a variational model to refine the depth.
Unlike these depth super-resolution methods, which take the dense and regular depth image as input.
Instead, the depth input in our method is sparse and irregular, and also we train our model end-to-end without any further optimization or post-processing.

{\bf Image-guided Methods}
These methods usually achieve better results, since they utilize an additional RGB image, which provides strong cues on semantic information, edge information, or surface information, etc.
Earlier works mainly address depth super-resolution with bilateral filtering~\cite{spatial_depth_range}, or global energy minimization~\cite{high_quality_3d_tof, high_quality_3d_tip, guided_depth_tgv},
where the depth completion is guided by image~\cite{spatial_depth_range, high_quality_3d_tof, high_quality_3d_tip, guided_depth_tgv}, semantic segmentation~\cite{semantic_upsampling} or edge information~\cite{edge_guided_depth}.

Recently, Zhang et al.~\cite{deep_indoor_rgbd} propose to predict surface normal and occlusion boundary from a deep network and further utilize them to help depth completion in indoor scenes.
Qiu et al.~\cite{deep_lidar} extend a similar surface normal as guidance idea to the outdoor environment and recover dense depth from sparse LiDAR data.
Ma et al.~\cite{self_supervised} propose a self-supervised network to explore photo-consistency among neighboring video frames for depth completion. 
Huang et al.~\cite{hms_net} propose three sparsity-invariant operations to deal with sparse inputs.
Eldesokey et al.~\cite{cnn_confidence} combine their confidence propagation~\cite{propagating_confidences} with RGB information to solve this problem.
Gansbeke et al.~\cite{sparse_noisy} use two parallel networks to predict depth and learn an uncertainty to fuse two results.
Cheng et al.~\cite{depth_affinity} use CNN to learn the affinity among neighboring pixels to help depth estimation.

Although various approaches have been proposed for depth completion with a reference RGB image,
they almost share the same strategy in fusing depth and image features, which is simple concatenation or element-wise addition operation.
In this paper, inspired by guided image filtering~\cite{guided_filter},
we propose a novel guided convolution module for feature fusion, to better utilize the guidance information from the RGB image. 

{\bf Joint Filtering and Guided Filtering}
Our method is also relevant to joint bilateral filtering~\cite{bilateral_filter} and guided image filtering~\cite{guided_filter}.
Joint/guided image filtering utilizes a reference or guidance image as prior and aims to transfer the structures from the reference image to the target image
for color/depth image super-resolution~\cite{joint_upsampling, spatial_depth_range}, image restoration~\cite{cross_restoration}, etc.

Early joint filtering methods~\cite{rolling_filter,mutual_filtering,robust_joint_filter} explore common structures between target and reference images
and formulate the problem as iterative energy minimization.  
Recently, Li et al.~\cite{deep_joint_filtering} propose a CNNs based joint filtering for image noise reduction, depth upsampling etc.,
but the joint filtering is implemented as a simple feature concatenation.
Gharbi et al.~\cite{deep_bilateral} generate affine parameters by a deep network to perform color transforms for image enhancement.
Lee et al.~\cite{depth_comp_geo_con} adopt a similar bilateral learning scheme of~\cite{deep_bilateral}
but generate bilateral weights and apply them once on a pre-obtained depth map for depth refinement. 
In contrast, our guided convolution module works on image features and serves as a flexibly pluggable component in multiple stages of an encoder-decoder network. 
 
In~\cite{fast_guide_filter}, Wu et al. propose a guided filtering layer to perform joint upsampling, which is close to our work.
It directly reformulates the conventional guided filter~\cite{guided_filter} and make it differentiable as a neural network layer.
As a result, the kernel weights are generated by the same close-form equation of guided filter~\cite{guided_filter} to filter the input image.
This kind of operator is inapplicable to fill-in sparse LiDAR points, as commented by the authors of guided filter in their conference paper~\cite{guided_filter_conf}.
Our method is also inspired by guided filter~\cite{guided_filter}. Rather than generating guided filter kernels from a specific close-form equation, 
we consider to learns more general and powerful kernels from the guidance image and applies the kernels to fuse multi-modal features for depth completion task.

{\bf Dynamic Filtering}
On the other hand, in convolutional neural networks, Dynamic Filtering Network (DFN)~\cite{dynamic_filter} is a broad category of methods
where the network generates filter kernels dynamically based on the input image to enable operations like local spatial transformation on the input features.
The general concept first proposed in~\cite{dynamic_filter} is mainly evaluated on video (and stereo) prediction with previous frames as input.

Recently, several applications and extensions of DFN have been developed.
`Deformable convolution'~\cite{deformable_conv} dynamically generates the offsets to the fixed geometric structure
which can be seen as an extension of DFN by focusing on the sampling locations.
Simonovsky et al.~\cite{edge_condition_filter} extends DFN into the graph signals in spatial domain,
where the filter weights are dynamically generated for each specific input sample and conditioned on the edge labels.
Wu et al.~\cite{large_dynamic_filter} propose an extension of DFN by using multiple sampled neighbor regions to dynamically generate weights with larger receptive fields.

Our kernel generating approach shares the same philosophy with DFN and can be considered as a variant and extension, focusing on multi-stage feature fusion of multi-modal data.    
The spatially-variant kernels generated by DFN~\cite{dynamic_filter} consume large GPU memories and thus are only applied once on low resolution images or features. 
However, multi-stage feature fusion is critical for feature extraction from sparse depth and color image on the depth completion task, but has not been studied by previous DFN papers.  
To address it, we design a novel network structure with convolution factorization and further discuss the impact of fusion strategies on depth completion results.

%!TEX root = root.tex

%------------------------------------------------------------------------
\section{The Proposed Method}
Given a sparse depth map $\mathbf{S}$ generated by projecting the LiDAR points to the image plane with calibration parameters and
a RGB image $\mathbf{I}$ as guidance reference,
depth completion aims to produce a dense depth map $\mathbf{D}$ of the whole image.
The RGB image can provide extremely useful information for depth completion task, as it depicts object boundaries and scene contents. 

To explain our guided convolutional network to upgrade $\mathbf{S}$ to $\mathbf{D}$ with the guidance of $\mathbf{I}$,
we first briefly review the guided image filtering which inspires our guided convolution module in subsection~\ref{subsec:guided_filtering}.
Then we elaborate the design of the guided convolution module in subsection~\ref{subsec:guide_module} and introduce a novel convolution factorization in subsection~\ref{subsec:conv_fact}.
In the next, we explain how this module can be used in a common encoder-decoder network, and the multi-stage fusion scheme used in our method in subsection~\ref{subsec:net_arch}.
Finally, we give implementation details including hyperparameter settings in subsection~\ref{subsec:implement_detail}.

\begin{figure*}
\begin{center}
\includegraphics[width=0.98\textwidth]{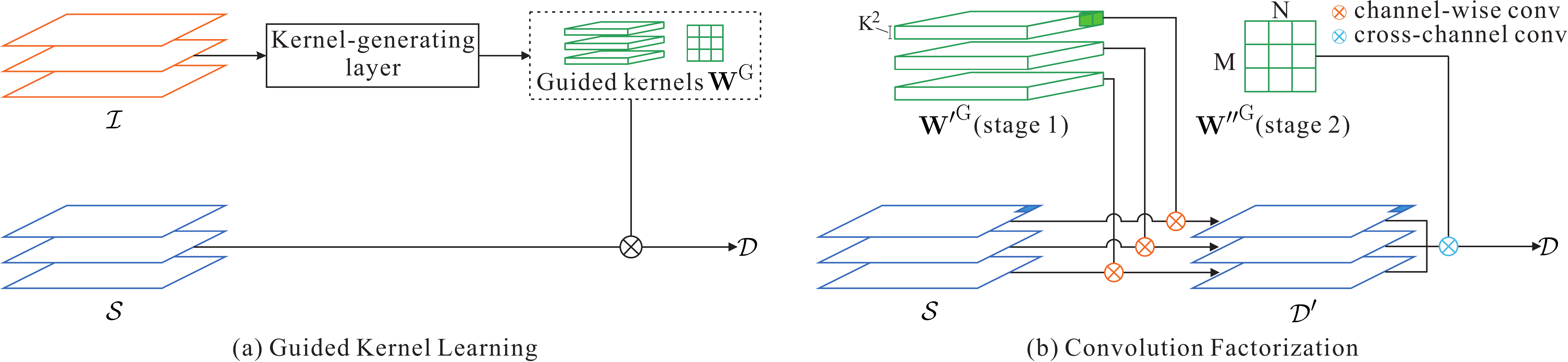}
\end{center}
   \caption{{\bf Guided Convolution Module.}
   (a) shows the overall pipeline of guided convolution module.
   Given image features $\mathcal{I}$ as input, filter generation layer dynamically produces guided kernels $\mathbf{W}^\mathrm{G}$
   (including $\mathbf{W'}^\mathrm{G}$ and $\mathbf{W''}^\mathrm{G}$),
   which are further applied on input depth features $\mathcal{S}$ and output new depth features $\mathcal{D}$.
   (b) shows the details of convolution between guided kernels $\mathbf{W}^\mathrm{G}$ and input depth features $\mathcal{S}$.
   We factorize it into two-stage convolutions: channel-wise convolution and cross-channel convolution.}
   \label{fig:guide_module}
\end{figure*}

\subsection{Guided Image Filtering}
\label{subsec:guided_filtering}
The guided image filtering\cite{guided_filter} generates \emph{spatially-variant} filters according to a guidance image.
In our setting of depth completion task, this method would compute the value at a pixel $i$ in $\mathbf{D}$ as a weighted average of nearby pixels from $\mathbf{S}$, i.e.
\begin{equation} \label{eq:guided_avg}
\mathbf{D}_i = \sum_{j \in \mathcal{N}(i)} \mathbf{W}_{ij}(\mathbf{I})\mathbf{S}_j.
\end{equation}
Here, $i$, $j$ are pixels indexes and $\mathcal{N}(i)$ is a local neighborhood of the pixel $i$.
The kernel weights $\mathbf{W}_{ij}$ are computed according to the guidance image $\mathbf{I}$ and a hand-crafted closed-form equation similar to the matting Laplacian from~\cite{close_form_matting}.
Unless specifically indicating, we omit the index of the image or feature channel for simplifying notations.

This guided image filtering might be applied to image super-resolution like in \cite{joint_geodesic_upsampling}.
However, our input LiDAR points are sparse and irregular. 
As pointed by the authors of~\cite{guided_filter_conf}, the guided image filtering cannot work well on sparse inputs.
This motivates us to learn more general and powerful filter kernels from the guidance image $\mathbf{I}$, rather than using the hand-crafted function for kernel generation.
And then we apply the kernels to fuse the multi-modal features, not directly filtering on the input images.

\subsection{Guided Convolution Module}
\label{subsec:guide_module}

Here, we elaborate the design of our guided convolution module that generates \emph{content-dependent} and \emph{spatially-variant} kernels for depth completion.

As shown in Figure~\ref{fig:net_arch},
our guided convolution module would server as a flexibly pluggable component to fuse the features from RGB and depth image in multiple stages.
It would generate convolutional kernels automatically from the guidance image feature $\mathcal{I}$ and apply them to the sparse depth map feature $\mathcal{S}$.
Here, $\mathcal{I}$ and $\mathcal{S}$ are features extracted from the guidance image $\mathbf{I}$ and sparse depth map $\mathbf{S}$ respectively.
We denote the output from this guided convolution module as $\mathcal{D}$, which is the extracted feature of depth image. Formally, 
\begin{equation}
\label{eq:our_guided_filter}
   \mathcal{D} = \mathbf{W}^{\mathrm{G}}(\mathcal{I}; \Theta) \otimes \mathcal{S},
\end{equation}
where $\mathbf{W}^{\mathrm{G}}$ is the kernel generated by our network according to the input guidance image feature $\mathcal{I}$, and further depends on the network parameter $\Theta$.
Here, $\otimes$ indicates the convolution operation. 

Figure~\ref{fig:guide_module} (a) illustrates the design of our learnable guided convolution module.
There is a \emph{`Kernel-Generating Layer' (KGL)} to generate the kernel $\mathbf{W}^\mathrm{G}$ according to the image features $\mathcal{I}$.
The parameters of the KGL are $\Theta$.
We can employ any differentiable operations for this KGL in principle.
Since we deal with grid images, convolution layers are preferable for this task.
Thus, the most na\"{\i}ve implementation is to directly apply convolution layers to generate all the kernel weights required for convolution operation on the depth feature map.
Please note that the kernel $\mathbf{W}^{\mathrm{G}}$ is  \emph{content-dependent} and \emph{spatially-variant}.
\emph{Content-dependent} means the guided kernel $\mathbf{W}^\mathrm{G}$ is dynamically generated, depending on the image content $\mathcal{I}$.
\emph{Spatially-variant} means different kernels are applied to different spatial positions of the sparse depth feature $\mathcal{S}$.
In comparison, $\Theta$ is fixed spatially and across different input images once it is learned.

The advantages of \emph{content-dependent} and \emph{spatially-variant} kernels are two-folds.
Firstly, this kind of kernels allows the network to apply different filters to different objects (and different image regions).
It is useful because, for example, the depth distribution on a car would be different from that on the road (also, nearby and faraway cars own different depth distributions).
Thus, generating the kernels dynamically according to the image content and spatial position would be helpful. 
Secondly, during training, the gradient of a spatially-invariant kernel is computed as the average over all image pixels from the next layer.
Such an average is more likely leading to gradient closing to zero, even thought the learned kernel is far from optimal for every position,
which could generate sub-optimal results as pointed by~\cite{large_dynamic_filter}.
In comparison, spatially variant kernels can alleviate this problem and make the training better behaved, which towards to stronger results.

\subsection{Convolution Factorization}
\label{subsec:conv_fact} 

However, generating and applying these spatially-variant kernels na\"{\i}vely would consume a large amount of GPU memory and computation resources.
The enormous GPU memory consumption is unaffordable for modern GPU device, 
when integrating the guided convolution module into multi-stage fusion of an encoder-decoder network.
To address this challenge, inspired by recent network compression techniques, e.g. MobileNets~\cite{MobileNets},
we design a novel factorization as well as a matched network structure to split the guided convolution module into two stages for better memory and computation efficiency.
This step is critical to make the network practical.

As shown in Figure~\ref{fig:guide_module} (b), the first stage is a \emph{channel-wise} convolution layer
where the $m$-th channel of the depth feature $\mathcal{S}_m$ is convolved with the corresponding channel of the generated filter kernel $\mathbf{W'}^\mathrm{G}_{m}$.
These convolutions are still spatially-variant. The output depth feature $\mathcal{D}'_m$ after the first stage then becomes
\begin{equation}
   \label{eq:conv_fact1}
      \mathcal{D}'_m = \mathbf{W'}^{\mathrm{G}}_m(\mathcal{I};\Theta') \otimes \mathcal{S}_m,
\end{equation}
where $\Theta'$ and $\mathbf{W'}^{\mathrm{G}}$ are the KGL parameter and the guided kernel in the first stage, respectively.
In this stage, the KGL is implemented by a standard convolution layer.

The second stage is a \emph{cross-channel} convolution layer where a $1 \times 1$ convolution aggregates features across different channels.
This stage is still content-dependent but spatially-invariant.
The kernel weights are also generated from the guidance image feature $\mathcal{I}$, but are shared among all pixels.
Specifically, we first use an average pooling over the guidance image feature $\mathcal{I}$ at each channel individually
to obtain an intermediate image feature $\mathcal{I}'$ with size $M \times 1 \times 1$, where $M$ is the number of channels of  $\mathcal{I}$.
We then feed $\mathcal{I}'$ into a fully-connected layer to generate the guided kernel $\mathbf{W''}^{\mathrm{G}}$, whose size is $M \times N \times 1 \times 1$, where $N$ is the number of channels of the dense depth feature $\mathcal{D}$.
Finally, we apply $\mathbf{W''}^{\mathrm{G}}$ to the depth feature $\mathcal{D}'$ from the \emph{channel-wise} convolution layer to obtain the final depth feature  $\mathcal{D}$. Formally, 
\begin{equation}
\label{eq:conv_fact2}
   \mathcal{D} = \mathbf{W''}^{\mathrm{G}}(\mathcal{I'};\Theta'' ) \otimes \mathcal{D'}, 
\end{equation}
where $\Theta''$ is the parameter in the fully-connected layer.
In Equation~(\ref{eq:conv_fact2}), $\mathbf{W''}^{\mathrm{G}}$ is spatially invariant and shared by all pixels.
The convolution applied to $\mathcal{D'}$ is a $1 \times 1$ convolution to aggregate this $M$-channel features to a $N$-channel features in $\mathcal{D}$ and can be executed quickly.

\textbf{Memory \& Computation Efficiency Analysis.}
Now, we analyze the improvement of this two-stage strategy in terms of memory and computation efficiency. 
If the convolution operations $\otimes$ in Equqation~(\ref{eq:our_guided_filter}) is implemented na\"{\i}vly,
the target depth feature $\mathcal{D}_{p,n}$ at a pixel $p$ and the channel $n$ can be formalized explicitly as 
\begin{equation}
\label{eq:standard_conv}
   \mathcal{D}_{p,n} = \sum_{m}\sum_{k}\mathbf{W}_{p,k,m,n}^{\mathrm{G}}(\mathcal{I};\Theta)\cdot \mathcal{S}_{p+k,m}, 
\end{equation}
where $k$ is the offset in a $K \times K$ filter kernel window centered at $p$ and $m$ is the channel index of $\mathcal{S}$.
Suppose the height and width of the input depth feature $\mathcal{S}$ are $H$ and $B$ respectively.
It is easy to figure out that the size of the generated kernel is $(M \times N \times K^2 \times H \times B)$.
In an encoder-decoder network, $H$ and $B$ are usually very large in the initial scales of  the encoder or the end scales of the decoder.
$M$ and $N$ usually go up to hundreds or even thousands in the latent space. Hence, the memory consumption is high and unaffordable even for modern GPUs. 

By our convolution factorization,
we split convolution in Equation~(\ref{eq:standard_conv}) into a \emph{channel-wise} convolution in Equation~(\ref{eq:conv_fact1})
and a \emph{cross-channel} convolution in Equation~(\ref{eq:conv_fact2}).
We can explicitly re-formulate these two equations in detail as
\begin{equation} \label{eq:exp_fact1} 
    \mathcal{D}'_{p,m} = \sum_{k}{\mathbf{W'}}_{p,k,m}^{\mathrm{G}}(\mathcal{I};\Theta') \cdot \mathcal{S}_{p+k,m} 
\end{equation}
 and
 \begin{equation} \label{eq:exp_fact2} 
    \mathcal{D}_{p,n} = \sum_{m}\mathbf{W''}_{m,n}^{\mathrm{G}}(\mathcal{I'};\Theta'') \cdot \mathcal{D'}_{p,m}, 
 \end{equation} 

The computation complexities of Equation~(\ref{eq:exp_fact1}) and Equation~(\ref{eq:exp_fact2}) are $O(K^2)$ and $O(M)$ respectively.
Therefore,
by this novel convolution factorization,
we reduce the computational complexity of $\mathcal{D}_{p,n}$ from $O(M \times K^2)$ to $O(M + K^2)$. 

Moreover, the proposed factorization can reduce GPU memory consumption enormously.
This is extremely important for networks with multi-stage fusions.
Suppose the memory consumption by the proposed factorization and na\"{\i}ve convolution are $M_f$ and $M_s$ respectively, then
\begin{equation}
\label{eq:memory_compare}
   \begin{split}
   & \frac{M_f}{M_s} = \frac{ M\times K^{2}\times H \times B + M \times N}{ M \times N\times K^{2}\times H \times B } \\
   & \quad \ \ \ =\frac{1}{N} + \frac{1}{K^{2}\times H \times B }.
   \end{split}
\end{equation}

As an example, when using 4-byte floating precision and taking $M = N = 128$, $H = 64$, $B=304$, and $K=3$,
which is the setting of the second fusion stage of our network, the proposed two-stage convolution reduces GPU memory from 10.7GB to 0.08GB, nearly 128 times lower for just a single layer.
In this way, our guided convolution module can be applied on multiple scales of a network, e.g. in an encoder-decoder network.

\subsection{Network Architecture}
\label{subsec:net_arch}
Figure~\ref{fig:net_arch} illustrates the overall structure of the proposed network, which is based on two encoder-decoder networks with skip layers.
Here, we refer the two networks taking the RGB image $\mathbf{I}$ and sparse LiDAR depth image $\mathbf{S}$ as \emph{GuideNet} and \emph{DepthNet} respectively.
The GuidedNet aims to learn hierarchical feature representations with both low-level and high-level information from RGB image.
Such image features are used to generate spatially-variant and content-dependent kernels automatically for depth feature extractions.
The DepthNet takes the LiDAR depth image as input and progressively fuse hierarchical image features by the guided convolution module in encoder stage.
It then regresses dense depth image at the decoder stage.
Both encoders of GuidedNet and DepthNet consist of a trail of ResNet blocks~\cite{res_net}.
Convolution layer with stride is used to aggregate feature to low resolution in encoder stage,
and deconvolution layer in decoder stage upsamples the feature map to high resolution.
We also add standard convolution layers at the beginning of both GuideNet and DepthNet as well as the end of DepthNet.  

Please note that during feature fusion, instead of the early or late fusion scheme widely used in the existing methods~\cite{deep_lidar, sparse_noisy, self_supervised},
we utilize a novel fusion scheme which fuse the \emph{decoder} features of the GuidedNet to the \emph{encoder} features of the DepthNet.
In our network, image features act as guidance for the generation of depth feature representations.
Thus, compared with encoder features, features from the decoder stage in the GuideNet are preferable, as they own more high-level context information.
In addition, in contrast to fuse only once,
we fuse the two sources in multi-stage, which shows stronger and more reliable results.
More comparisons and analyses can be found in subsection~\ref{subsec:ablation_studies}.

\subsection{Implementation Details}
\label{subsec:implement_detail}

\subsubsection{Loss Function}
During training, we adopt the mean squared error (MSE) to compute the loss between ground truth and predicted depth.
For real-world data, the ground truth depth is often semi-dense, because it is difficult to collect ground truth depth for every pixel.
Therefore, we only consider valid pixels in the reference ground truth depth map when computing the training loss.
The final loss function is
\begin{equation} \label{eq:loss}
L = \sum_{p \in \mathbf{P}_{v}} \|\mathbf{D}_p^\mathrm{gt}-\mathbf{D}_p\|^{2},
\end{equation}
where \(\mathbf{P}_{v}\) represents the set of valid pixels.
$\mathbf{D}^\mathrm{gt}_p$ and $\mathbf{D}_p$ denote the ground truth and predicted depth at the pixel~$p$, respectively.

\subsubsection{Training Setting}
 We use ADAM\cite{adam} as the optimizer with a starting learning rate of $10^{-3}$ and weight decay of $10^{-6}$.
 The learning rate drops by half every $50k$ iterations.
 We utilize 2 GTX 1080Ti GPUs for training with batch size of 8.
 Synchronized Cross-GPU Batch Normalization~\cite{batch_norm,pytorch_encoding} is used in the network training stage.
 Our method is trained end-to-end FROM SCRATCH.
 In contrast, some state-of-the-art methods employ extra datasets for training,
 e.g. DeepLiDAR~\cite{deep_lidar} utilizes synthetic data to train the network for obtaining scene surface normal,
 and the authors of~\cite{sparse_noisy} use a pretrained model on Cityscapes\footnote{\url{https://www.cityscapes-dataset.com}} as network initialization.

%!TEX root = root.tex

%------------------------------------------------------------------------
\section{Experiments}

We conduct comprehensive experiments to verify our method on both outdoor and indoor datasets, captured in real-world and synthetic scenes.
We first introduce all the datasets and evaluation metrics used in our experiments in subsection~\ref{subsec:dataset} and~\ref{subsec:metric} respectively.
Then, as autonomous driving is the major application of depth completion,
we compare our method with the state-of-the-art methods on the outdoor scene KITTI dataset in subsection~\ref{subsec:result_kitti}.
It follows by extensive ablation studies on the KITTI validation set in subsection~\ref{subsec:ablation_studies} to investigate the impact of each network component and the fusion scheme used in our method.
In subsection~\ref{subsec:result_nyu_v2}, we verify the performance of proposed method on the indoor scene NYUv2 dataset.
Finally, in subsection~\ref{subsec:generalization_capability}, we perform experiments under various settings including input depth with different densities,
RGB images captured under various lighting and weather conditions and cross-dataset evaluations to prove generalization capability of our method.

\subsection{Datasets}
\label{subsec:dataset}

{\bf KITTI Dataset} The KITTI depth completion dataset~\cite{sparsity_cnn} contains $86,898$ frames for training, $1,000$ frames for validation, and another $1,000$ frames for testing.
It provides public leaderboard\footnote{\url{http://www.cvlibs.net/datasets/kitti/eval_depth.php?benchmark}} for ranking submissions. 
The ground truth depth is generated by registering LiDAR scans temporally.
These registered points are further verified with the stereo image pairs to get rid of noisy points.
As there are rare LiDAR points at the top of an image, following~\cite{sparse_noisy}, input images are cropped to \(256 \times 1216\) for both training and testing.

{\bf Virtual KITTI Dataset} Virtual KITTI dataset~\cite{vkitti} is a synthetic dataset, where the virtual scenes are cloned from the real world
KITTI video sequences. 
Besides the 5 virtual image sequences cloned from KITTI sequence,
it also generates the corresponding image sequences under various lighting conditions (like morning, sunset) and weather conditions (like fog, rain),
totally 17,000 image frames.
To generate sparse LiDAR points, instead of random sampling from the dense depth map,
we use the sparse depth of the corresponding image frame in KITTI dataset as a mask to obtain sparse samples from dense ground truth depth,
which makes the distribution of sparse depth on image is close to real-world situation.
We split the whole Virtual KITTI dataset to train and test set to fine-tune and evaluate our model respectively.
Since the destination is to verify the robustness of our model under various lighting and weather condition,
we only fine-tune our model under the original `clone' condition whose weather is good, using sequence of `0001', `0002', `0006' and `0018' for training.
And the sequence `0020' with various weather and lighting conditions is used for evaluation. 
In summary, we have 1289 frames for fine-tuning and 837 frames for each condition to evaluate.

\begin{figure*}
   \begin{center}
   \includegraphics[width=0.98\textwidth]{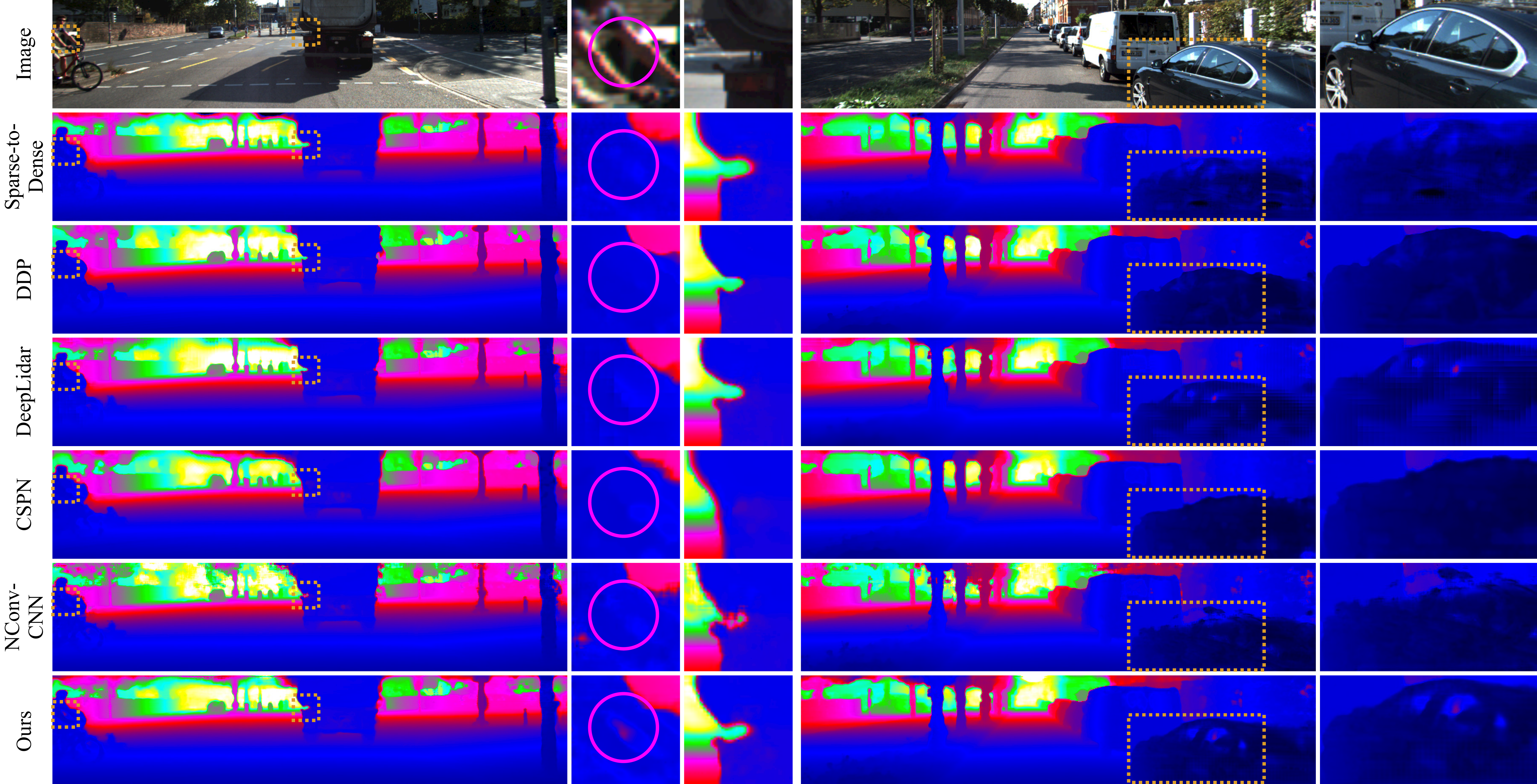}
   \end{center}
      \caption{Qualitative comparison with state-of-the-art methods on KITTI test set.
      The results are from the KITTI depth completion leaderboard in which depth images are colorized along with depth range.
      Our results are shown in the bottom row and compared with top-ranking methods
      `Sparse-to-Dense'~\cite{sparse_to_dense}, `DDP'~\cite{ddp}, `DeepLiDAR'~\cite{deep_lidar}, `CSPN'~\cite{depth_affinity,cspn_journal} and `NConv-CNN'~\cite{cnn_confidence}.
      In the zoomed regions, our method recovers better 3D details.
      }
   \label{fig:compare_kitti}
\end{figure*}

{\bf NYUv2 Dataset}
NYUv2 dataset~\cite{NYU_V2} consists of RGB images and depth images captured by Microsoft Kinect in 464 indoor scenes. 
Following the similar setting of previous depth completion methods~\cite{depth_affinity,deep_lidar,sparse_to_dense}, 
our method is trained on $50k$ images uniformly sampled from the training set, and tested on the 654 official labeled test set for evaluation.
As a preprocessing, the depth values are in-painted using the official toolbox, which adopts the colorization scheme~\cite{colorization} to fill-in missing values.
For both train and test set, the original frames of size $640\times480$ are half down-sampled with bilinear interpolation, and then center-cropped to $304\times228$. 
The sparse input depth is generated by random sampling from the dense ground truth.
Due to the input resolution for our network must be a multiple of $32$,
we futher pad the images to $320\times256$ as input for our method but evaluate only the valid region of size $304\times228$ to keep fair comparison with other methods.

{\bf SUN RGBD Dataset}
The SUN RGBD dataset~\cite{sun_rgbd} is an indoor dataset containing RGB-D images from many other datasets~\cite{NYU_V2,B3DO,sun_3d}.
We only use SUN RGBD dataset for cross-dataset evaluation.
Since NYUv2 dataset is a subset of SUN RGBD dataset, we exclude them in evaluation to avoid repetition.
We keep all images with the same resolution of NYUv2 dataset as $640\times480$, captured under different scenes.
Totally, we evaluate our model on 3944 image frames, with 555 frames captured by Kinect V1 and 3389 captured by Asus Xtion camera.
The same pre-processing method for NYUv2 dataset is used to fill depth map.
Note, frames captured by Asus Xtion camera are more challenging, because the data comes from a different device.

\subsection{Evaluation Metrics} 
\label{subsec:metric}
Following the KITTI benchmark and exiting depth completion methods~\cite{depth_affinity, deep_lidar, self_supervised}, for outdoor scene,
we use these four standard metrics for evaluation:
root mean squared error (RMSE), mean absolute error (MAE), root mean squared error of the inverse depth (iRMSE) and mean absolute error of the inverse depth (iMAE).
Among them, RMSE and MAE directly measure depth accuracy, while RMSE is more sensitive and chosen as the dominant metric to rank submissions on the KITTI leaderboard. 
iRMSE and iMAE compute the mean error of inverse depth, which gives less weight for far-away points.

For indoor scene, to be consistent with comparative depth completion methods~\cite{depth_affinity, deep_lidar, self_supervised, cnn_confidence},
the evaluation metrics are selected as root mean squared error (RMSE), mean absolute relative error (REL) and
$\delta_{i}$ which means the percentage of predicted pixels where the relative error is less a threshold $i$.
Specifically, $i$ is chosen as $1.25$, $1.25^{2}$ and $1.25^{3}$ separately for evaluation.
Here, a higher $i$ indicates a softer constraint and a higher $\delta_{i}$ represents a better prediction.
RMSE is chosen as the primary metric for all the experiment evaluations as it is sensitive to large errors on distant regions.

\subsection{Experiments on KITTI Dataset}
\label{subsec:result_kitti}

We first evaluate our method on the KITTI depth completion dataset~\cite{sparsity_cnn}.
Our method is trained end-to-end from scratch on the train set and compared the performance with state-of-the-art methods on test set. 
Table~\ref{tab:kitti_result} lists the quantitative comparison of our method and other top-ranking published methods on the  KITTI leaderboard.
Our method ranks 1st and exceed all other methods under the primary RMSE metric at the time of paper submission,
and presents comparable performance on other evaluation metrics.

Figure~\ref{fig:compare_kitti} shows some visual comparison results with several state-of-the-art methods on the KITTI test set.
Our results are shown in the last row.
While all methods provide visually plausible results in general, our estimated depth maps reveal more details and are more accurate around object boundaries.
For example, our method can better recover depth of background between the arms of a person as highlighted by the magenta circle in Figure~\ref{fig:compare_kitti}.
The predicted depth of our method owns the most accurate contour in the black car region. 

\begin{figure*}
   \begin{center}
   \includegraphics[width=0.98\textwidth]{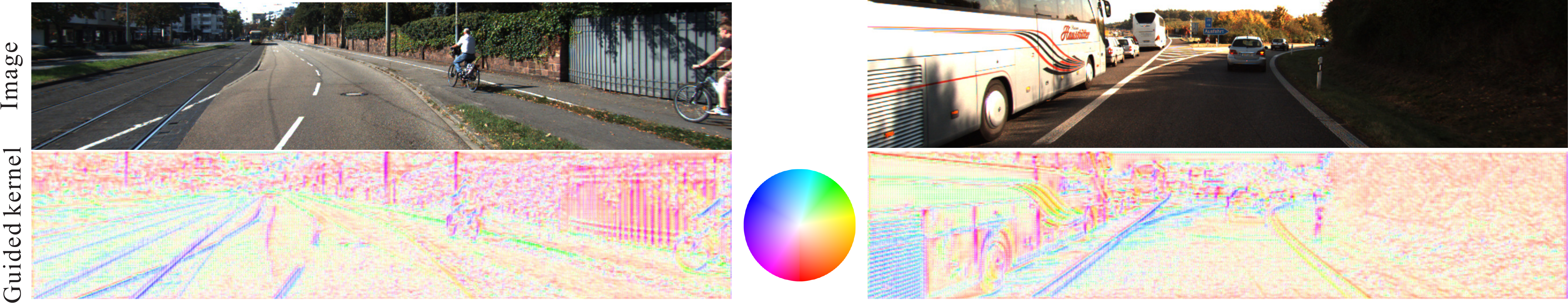}
   \end{center}
      \caption{Visualization of the guided kernels, where a kernel is visualized as a 2D vector by applying the Prewitt operator~\cite{prewitt}. Similar pixels tend to have the similar kernels.}
   \label{fig:flow}
\end{figure*}

\begin{table}
   \begin{center}
   \caption{{\bf Performance on the KITTI dataset.} The result is evaluated by the KITTI testing server and different methods are ranked by the RMSE (in $mm$).} \label{tab:kitti_result}
   \small
   \begin{tabular}{l|c|c|c|c} \hline
    & \textbf{RMSE}   & MAE   & iRMSE & iMAE \\ \hline
   CSPN~\cite{depth_affinity,cspn_journal} & 1019.64 & 279.46 & 2.93  & 1.15 \\
   DDP~\cite{ddp} & 832.94 & \textbf{203.96} & \textbf{2.10} & \textbf{0.85} \\
   NConv-CNN~\cite{cnn_confidence} & 829.98 & 233.26 & 2.60 & 1.03 \\
   Sparse-to-Dense~\cite{self_supervised} & 814.73 & 249.95 & 2.80 & 1.21 \\
   RGB\_certainty~\cite{sparse_noisy} & 772.87 & 215.02 & 2.19  & 0.93 \\
   DeepLiDAR~\cite{deep_lidar} & 758.38 & 226.50 & 2.56 & 1.15 \\  \hline
   Ours & \textbf{736.24} & 218.83 & 2.25 & 0.99 \\ \hline
   \end{tabular}
   \end{center}
\end{table}

Furthermore, to verify whether the guided convolution module really learns content-dependent and spatially-variant information to benefit depth completion,
we visualize one selected channel of the guided kernels $\mathbf{W'}^{\mathrm{G}}$ from the most early fusion stage in Figure~\ref{fig:flow}. 
This is done by applying the Prewitt operator~\cite{prewitt} on each $K \times K$ kernel to get the weighted sum of $x$-axis shift and $y$-axis shift, respectively.
We then obtain a 2D vector at each pixel and visualize it by a color code, like the way optical flow is visualized.
We can easily see that the boundary with similar gradient direction or surface with similar normal direction share similar color code. 
Please note that the method used here to visualize the guided kernels is extremely rough due to the difficulty of kernel weight interpretation in deep neural networks.
Also, the network is only supervised by semi-dense depth,
it's almost impossible for each object has it own color code in visualization without direct semantic supervision,
as semantic information is defined by human beings and owns little relationship with the depth supervision.  
To some extent, this visualization confirms the guided kernels are consistent with image content.
Hence the guided kernels are likely helpful for depth completion.

\subsection{Ablation Studies}
\label{subsec:ablation_studies}
To investigate the impact of each network component and fusion scheme on the final performance, we conduct ablation studies on the KITTI validation dataset.
Specifically, we evaluate several different variations of our network.
The quantitative comparisons are summarized in Table~\ref{tab:ablation}.

\subsubsection{Comparison with Feature Addition/Concatenation}
Existing methods often use addition or concatenation for multi-modality feature fusion.
To compare with them, we replace all the guided convolution modules in our network by feature addition or concatenation but keep the other network components and settings unchanged.
The results are indicated as `\texttt{Add.}' and `\texttt{Concat.}' respectively.
Compared with our guided convolution module, the simple feature addition or concatenation significantly worsen the results,
with the RMSE increasing 31.59 $\mathrm{mm}$ and 24.35 $\mathrm{mm}$ respectively. 

We can see that the results of `\texttt{Add.}' is a slightly worse than that of `\texttt{Concat.}'.
This is also reasonable, because image and depth features are heterogeneous data from different sources. 
By applying addition, we implicitly treat these two different features in the same way, which leads to performance drops.
Indeed, most of state-of-the-art methods~\cite{deep_lidar,self_supervised,cnn_confidence} adopt concatenation to
fuse the heterogeneous depth and image features while apply addition to fuse homogeneous depth features from different stages.

\subsubsection{Fusion Scheme of GuideNet and DepthNet}
As described in subsection~\ref{subsec:net_arch}, instead of using early or late feature fusion like existing methods~\cite{deep_lidar, sparse_noisy, self_supervised},
our approach fuses the \emph{decoder} features of the GuideNet to the \emph{encoder} features of the DepthNet.
To verify the effectiveness of such a fusion scheme,
we train and evaluate the performance of fusing the decoder features of the GuideNet to the decoder features of the DepthNet (referred as `\texttt{D-D Fusion}')
and fusing the encoder features of the GuideNet to the encoder features of the DepthNet (referred as `\texttt{E-E Fusion}').
In the later one, the decoder structure of the GuideNet is removed since it is not used anymore.
In this way, our method can be seen as `\texttt{D-E Fusion}'.

\begin{table}
   \caption{{\bf Ablation study on KITTI's validation set.} See text in subsection~\ref{subsec:ablation_studies} for more details.} \label{tab:ablation} 
   \begin{center}
   \small
   \begin{tabular}{l|c|c|c|c} \hline
    & \textbf{RMSE}   & MAE   & iRMSE & iMAE \\ \hline
   \texttt{Add.} & 809.37 & 233.18 & 3.98 & 1.11 \\
   \texttt{Concat.} & 802.13 & 226.87 & 2.53 & 1.02 \\
   \texttt{E-E Fusion} & 783.35 & 222.43 & 2.51 & 1.01 \\
   \texttt{D-D Fusion} & 795.64 & 223.95 & 6.73 & 1.15 \\
   \texttt{First Guide} & 799.03 & 224.27 & 2.66 & 1.01 \\
   \texttt{Last Guide} & 800.60 & 226.07 & 2.68 & 1.03 \\  \hline
   Ours & \textbf{777.78} & \textbf{221.59} & \textbf{2.39} & \textbf{1.00} \\ \hline
   \end{tabular}
   \end{center}
   
\end{table}

Table~\ref{tab:ablation} compares the results of `\texttt{E-E Fusion}' and `\texttt{D-D Fusion}' with our method.
The performance drop of the `\texttt{E-E Fusion}' verifies our earlier analysis that the decoder image features own more high-level context information thus can better guide depth feature extraction. 
The `\texttt{D-D Fusion}', fusing image and depth features in the decoder stage, suffers from even larger performance drop.
Comparing the `\texttt{D-D Fusion}' and our final model, we conclude that the image guidance is more effective at encoder stage of depth feature extraction.
It's also reasonable and easy to understand, as feature extracted in early stage can influence the following feature extraction, especially for sparse depth image.

\begin{figure*}
   \begin{center}
   \includegraphics[width=0.98\textwidth]{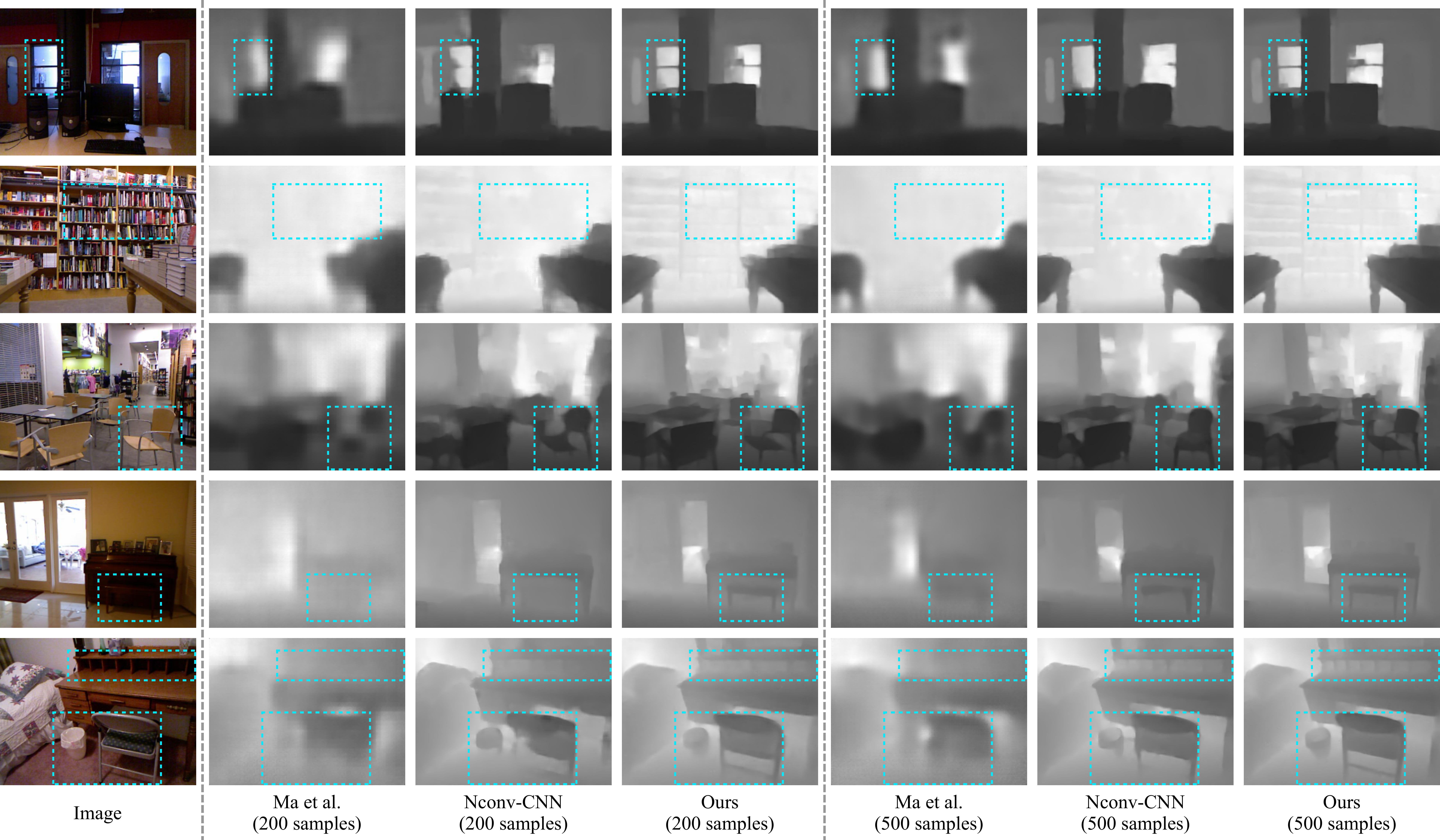}
   \end{center}
      \caption{Qualitative comparison with `Ma et al.'~\cite{sparse_to_dense} and `NConv-CNN'~\cite{cnn_confidence} on NYUv2 test set.
      We present the results of these three methods under 200 samples and 500 samples.
      Depth images are showed as grey images for clear visualization.
      The most notable regions are selected with cyan rectangles for easy comparisons.
      }
   \label{fig:compare_nyu_v2}
\end{figure*}

On the other hand, even the weaker fusion strategy in the `\texttt{E-E Fusion}' outperforms conventional feature addition or concatenation.
This attributes to our guided convolution module that can generate content-dependent and spatially-variant kernels to promote the depth completion.
This observation further proves the effectiveness of the proposed guided convolution module.

\subsubsection{Fusion Scheme of Multi-stage Guidance}
We also design two other variants to verify the effectiveness of multi-stage guidance scheme.
For comparison, based on our guided network, we replace all the guided modules with concatenation except the one in the first fusion stage,
and refer it as `\texttt{First Guide}'.
From the same view, we use `\texttt{Last Guide}' to refer the condition only the guided module in the last fusion stage is remained.
Using concatenation for the feature fusion of other stages is from the result,
that concatenation can perform a little better than addition operation as shown in Table~\ref{tab:ablation}.

We can see that both the results of `\texttt{First Guide}' and `\texttt{Last Guide}' are worse than our multi-stage guidance scheme.
This demonstrates the effectiveness of our multi-stage guidance design.
Also, the `\texttt{First Guide}' performs a little bit better than `\texttt{Last Guide}'.
It also consists with our early analysis that image guidance is more effective at early stage,
since feature extracted in early stage can influence the following feature extraction.
Moreover, both the results of `\texttt{First Guide}' and `\texttt{Last Guide}' perform better than the `\texttt{Concat.}'.
It once more verifies that the designed Guided Convolution Module is a much powerful fusion scheme for depth completion.

\subsection{Experiments on NYUv2 Dataset}
\label{subsec:result_nyu_v2}
To verify the performance of our method on indoor scene, 
we directly train and evaluate our guided network on the NYUv2 dataset~\cite{NYU_V2}, without any specific modification. %a small dataset captured by ourselves.

\begin{table}
   
   \caption{{\bf Performance on the NYUv2 dataset.} Both settings of 200 samples and 500 samples are evaluated.} \label{tab:nyu_v2_result}
   \begin{center}
   \scalebox{0.85}{
   \begin{tabular}{c|l|c|c|c|c|c} \hline
            samples            & method & \textbf{RMSE}$\downarrow$   & REL$\downarrow$  & $\delta_{1.25}$$\uparrow$ & $\delta_{1.25^{2}}$$\uparrow$ & $\delta_{1.25^{3}}$$\uparrow$ \\ \hline
                              & Bilateral~\cite{NYU_V2} & 0.479 & 0.084 & 92.4 & 97.6 & 98.9 \\
                              & TGV~\cite{guided_depth_tgv} &  0.635 & 0.123 & 81.9 & 93.0 & 96.8 \\
                              & Zhang et al.~\cite{deep_indoor_rgbd} & 0.228 & 0.042 & 97.1 & 99.3 & 99.7 \\
            500              & Ma et al.~\cite{sparse_to_dense} & 0.204 & 0.043 & 97.8 & 99.6 & 99.9 \\
                              & NConv-CNN~\cite{cnn_confidence} & 0.129 & 0.018 & 99.0 & 99.8 & \textbf{100} \\
                              & CSPN~\cite{depth_affinity} & 0.117&0.016&99.2&\textbf{99.9}&\textbf{100}\\
                              & DeepLiDAR~\cite{deep_lidar} &  0.115&0.022&99.3&\textbf{99.9}&\textbf{100}\\ \cline{2-7}
                              & Ours &  \textbf{0.101}&\textbf{0.015}&\textbf{99.5}&\textbf{99.9}&\textbf{100}\\ \hline
                              & Ma et al.~\cite{sparse_to_dense}&  0.230&0.044&97.1&99.4&99.8 \\ 
            200                & NConv-CNN~\cite{cnn_confidence} &  0.173&0.027&98.2&99.6&99.9\\ \cline{2-7}
                              & Ours &  \textbf{0.142}&\textbf{0.024}&\textbf{98.8}&\textbf{99.8}&\textbf{100} \\ \hline
   \end{tabular}
   }
   \end{center}
\end{table}

Following existing methods, we train and evaluate our method with the settings of 200 and 500 sparse LiDAR samples separately.
The quantitative comparisons with other methods are shown in Table~\ref{tab:nyu_v2_result}.
The results of `Bilateral'~\cite{NYU_V2}, and `CSPN'~\cite{depth_affinity} come from the CSPN~\cite{depth_affinity}.
The results of `TGV'~\cite{guided_depth_tgv}, `Zhang et al.'~\cite{deep_indoor_rgbd} and `DeepLiDAR'~\cite{deep_lidar} are obtained from DeepLiDAR~\cite{deep_lidar}.
By using the released implementations, we get the results of `Ma et al.'~\cite{sparse_to_dense} with 500 samples and `NConv-CNN'~\cite{cnn_confidence} with 200 samples.
We can see from the results, our method outperforms all other methods in both settings of 500 samples and 200 samples.
Without specific modification, our method ranks top under all these 5 evaluation metrics.

We also show some qualitative comparisons on the test set in Figure~\ref{fig:compare_nyu_v2}.
Our method is compared with `NConv-CNN'~\cite{cnn_confidence} and `Ma et al.'~\cite{sparse_to_dense} on the settings of 200 samples and 500 samples.
The most notable regions are selected with cyan rectangles for easy comparisons.
From the predicted depth, we can see the results of `Ma et al.' over-smooth the whole image and blur small objects. 
Even though `NConv-CNN' shows much clear depth predictions, it also suffers obvious detail loss at object structures, especially the thin object boundaries.
Our method show sharp transitions aligning to local details and generate the best results.

\subsection{Generalization Capability}
\label{subsec:generalization_capability}
To prove the generalization capability of our method,
we test its performance under different point densities, various lighting and weather conditions as well as cross-dataset evaluations.

\subsubsection{Different Point Densities}
We test the performance of our method under different point densities.
Our model is the same one trained from scratch only on the KITTI train set without any fine-tuning, to faithfully reflect its generalization capability.
For a comparison, we also evaluate another two state-of-the-art methods, `NConv-CNN'~\cite{cnn_confidence} and `Sparse-to-Dense'~\cite{self_supervised},
using their open-source code and the best performed model trained by their authors.

\begin{figure}
   \begin{center}
   \includegraphics[width=0.98\columnwidth]{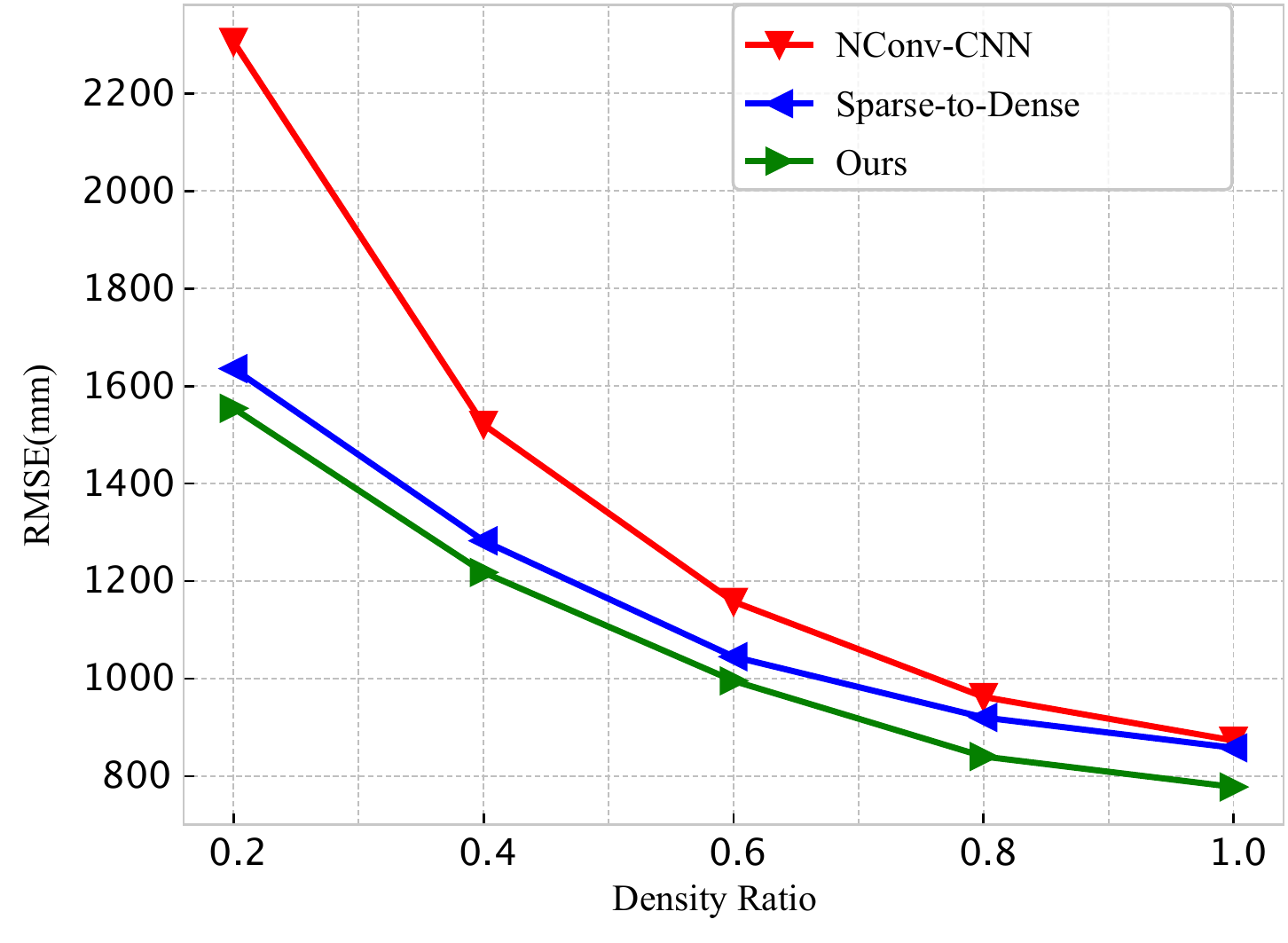}
   \end{center}
      \caption{
      RMSE (in $\mathrm{mm}$) under different levels of input LiDAR point density. The performance of our method and the `Sparse-to-Dense'~\cite{self_supervised} degrades more gently comparing to that of `NConv-CNN'~\cite{cnn_confidence}. }
   \label{fig:density_compare}
\end{figure}

Firstly, we vary the LiDAR input with 5 different levels of density on the KITTI validation set. 
The KITTI dataset is captured with a 64-line Velodyne LiDAR.
However, real industrial applications may only adopt a 32-line or even 16-line LiDAR considering the high sensor cost.
To analyze the impact of the sparsity level on the final result, we test with 5 different levels of LiDAR density on the KITTI validation dataset,
where the input LiDAR points are randomly sampled according to a given ratio.
Specifically, the density ratios of $0.2$, $0.4$, $0.6$, $0.8$ and $1.0$ are adopted in our evaluation.

Figure~\ref{fig:density_compare} shows the RMSE of our network, `NConv-CNN'~\cite{cnn_confidence} and `Sparse-to-Dense'~\cite{self_supervised} under various LiDAR point densities.
With the density decreasing, the `NConv-CNN'~\cite{cnn_confidence} shows significant performance drop and its RMSE increases quickly.
In comparison, our method and the `Sparse-to-Dense'~\cite{self_supervised}, on the other hand, degrade gradually and are consistently better than the `NConv-CNN'~\cite{cnn_confidence}.
The results demonstrate the strong generalization capability of our method under various LiDAR points density ratios.

\subsubsection{Various Lighting and Weather Conditions}
KITTI dataset is collected in the similar lighting condition and in good weather condition.
However, varied weather and lighting conditions always occur in practice and may bring the potential impact on the performance of depth completion.
To verify whether our guided network can still work well in these kinds of challenging situations,
we conduct evaluation experiments on Virtual KITTI dataset~\cite{vkitti} with various lighting (e.g., sunset) and weather (e.g., fog) conditions,
and compare our method with other two variants of `\texttt{Add.}' and `\texttt{Concat}' introduced in subsection~\ref{subsec:ablation_studies}.
Based on the trained model on KITTI dataset, we fine-tune our method under good `clone' condition, then test its performance under various lighting and weather condition in a different sequence.

\begin{figure}
   \begin{center}
   \includegraphics[width=0.98\columnwidth]{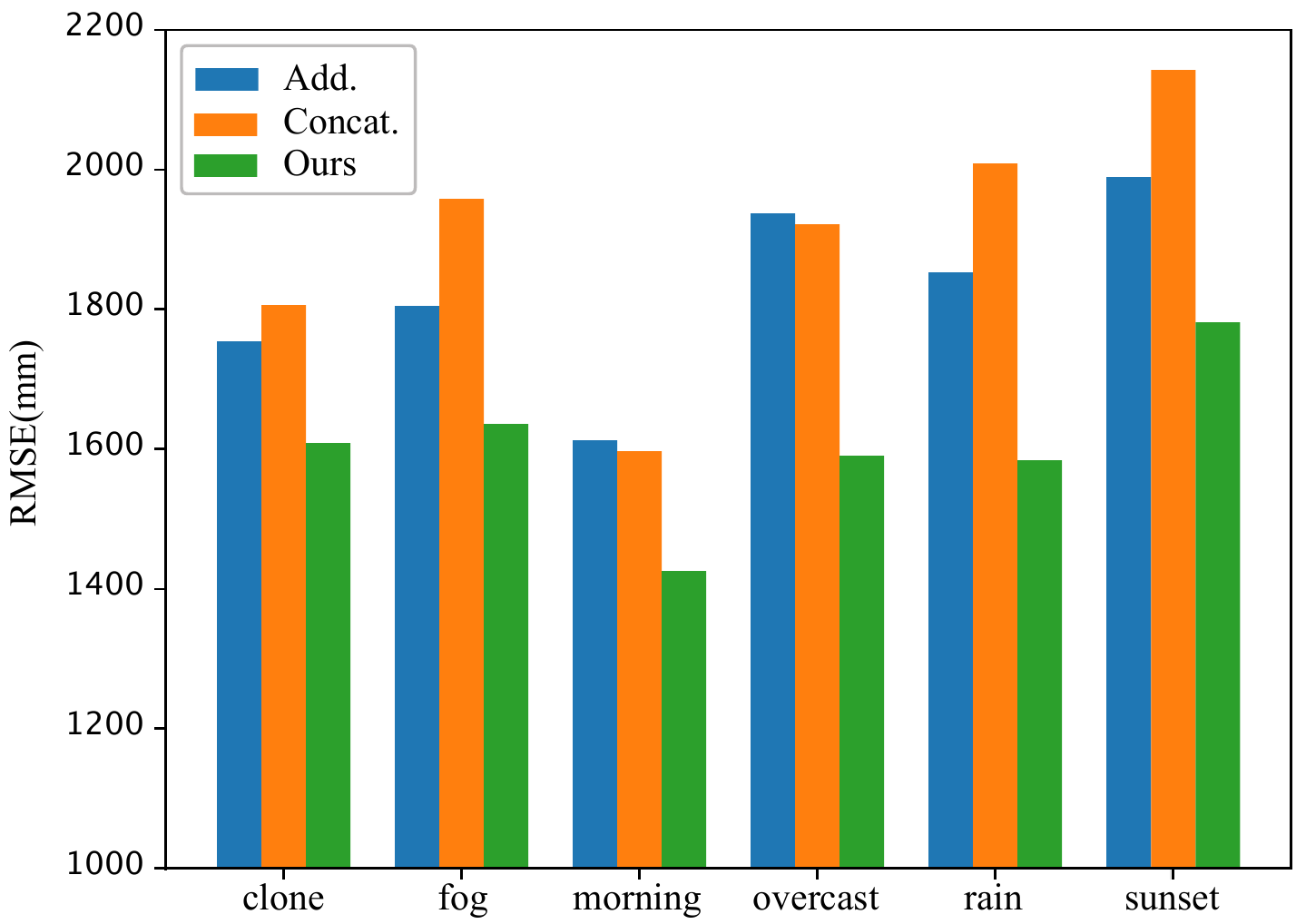}
   \end{center}
      \caption{RMSE (in $\mathrm{mm}$) on Virtual KITTI test set under various lighting and weather conditions.
      Our guided network are compared with the `\texttt{Add.}' and `\texttt{Concat}' variants.}
   \label{fig:compare_vkitti}
\end{figure}

\begin{figure*}
   \begin{center}
   \includegraphics[width=0.98\textwidth]{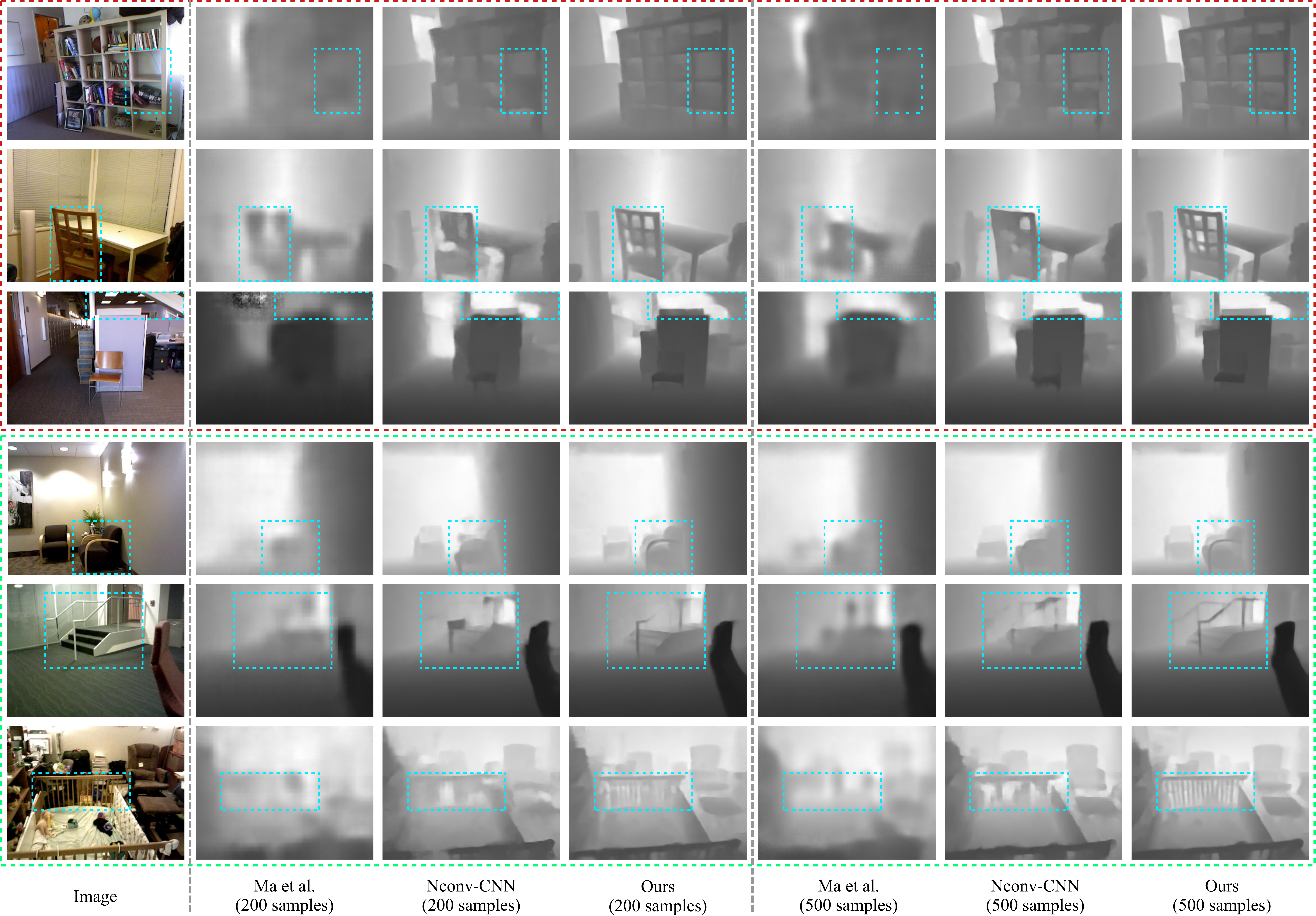}
   \end{center}
      \caption{Qualitative comparison with `Ma et al.'~\cite{sparse_to_dense} and `NConv-CNN'~\cite{cnn_confidence} on SUN RGBD dataset.
      Images in red rectangle are captured by Kinect V1 and Images in green rectangle are collected by Xtion.
      Depth results of these three methods under 200 samples and 500 samples are showed as grey images for clear visualization.
      The most notable regions are selected with cyan rectangles for easy comparisons.}
   \label{fig:compare_sunrgbd}
\end{figure*}

We evaluate our methods and two variants under the `clone', `fog', `morning', `overcast', `rain' and `sunset' conditions separately.
Figure~\ref{fig:compare_vkitti} depicts the results of three methods under various conditions.
We can easily find, compared with `\texttt{Add.}' and `\texttt{Concat}', our method achieves the best RMSE among all the conditions.
Also, the RMSE results of our method keep stable across all the situations,
which can verify the generalization capability of our method under various lighting and weather conditions.

\subsubsection{Cross-dataset Evaluation}
In order to show the generalization of our method,
we also conduct cross-dataset evaluations by using the models trained on NYUv2 dataset to directly test on SUN RGBD dataset~\cite{sun_rgbd}.

The comparison results are listed in Table~\ref{tab:B3DO} and Table~\ref{tab:sun_3d} for dataset captured by Kinect V1 and Asus Xtion camera respectively.
Both settings of 500 samples and 200 samples are evaluated by using the comparison models trained on NYUv2 dataset.
We can see our method still outperforms other methods with the best RMSE and reports close results with NYUv2 dataset.
The results demonstrate the strong cross-dataset generalization capability of our method.
We also present some quantitative results in Figure~\ref{fig:compare_sunrgbd}.
The first three rows selected in red rectangle are results on images captured by Kinect V1,
and the last three rows in green rectangle are results from Xtion.
The priority of our method can be found easily from the predicted depth, especially the selected regions.

\begin{table}
   
   \caption{{\bf Performance on the SUN RGBD dataset collected by Kinect V1.}  The evaluation frames are captured with same device as NYUv2 dataset.} \label{tab:B3DO}
   \begin{center}
   \scalebox{0.85}{
   \begin{tabular}{c|l|c|c|c|c|c} \hline
            samples            & method & \textbf{RMSE}$\downarrow$   & REL$\downarrow$  & $\delta_{1.25}$$\uparrow$ & $\delta_{1.25^{2}}$$\uparrow$ & $\delta_{1.25^{3}}$$\uparrow$ \\ \hline
                              & Ma et al.~\cite{sparse_to_dense}& 0.180 & 0.053 & 97.0 & 99.3 & 99.7 \\
            500              & Nconv-CNN~\cite{cnn_confidence}   & 0.119 & \textbf{0.019} & 98.7 & 99.7 & \textbf{99.9} \\ \cline{2-7}
                              & Ours &  \textbf{0.096}& 0.020&\textbf{99.0}&\textbf{99.8}&\textbf{99.9}\\ \hline
                              & Ma et al.~\cite{sparse_to_dense}&  0.206&0.044&97.1&99.4&99.8 \\ 
            200                & Nconv-CNN~\cite{cnn_confidence} &  0.159&\textbf{0.029}&\textbf{97.8}&99.4&99.8\\ \cline{2-7}
                              & Ours &  \textbf{0.139}&0.036&97.6&\textbf{99.5}&\textbf{99.9} \\ \hline
   \end{tabular}
   }
   \end{center}
\end{table}

\begin{table}
   
   \caption{{\bf Performance on the SUN RGBD dataset collected by Xtion.} The evaluation frames are captured with different device from NYUv2 dataset.} \label{tab:sun_3d}
   \begin{center}
   \scalebox{0.85}{
   \begin{tabular}{c|l|c|c|c|c|c} \hline
            samples            & method & \textbf{RMSE}$\downarrow$   & REL$\downarrow$  & $\delta_{1.25}$$\uparrow$ & $\delta_{1.25^{2}}$$\uparrow$ & $\delta_{1.25^{3}}$$\uparrow$ \\ \hline
                              & Ma et al.~\cite{sparse_to_dense}& 0.206 & 0.050 & 97.0 & 99.3 & 99.8 \\
            500              & Nconv-CNN~\cite{cnn_confidence}   & 0.136 & \textbf{0.020} & 98.6 & 99.6 & \textbf{99.9} \\ \cline{2-7}
                              & Ours &  \textbf{0.119}&\textbf{0.020}&\textbf{98.9}&\textbf{99.8}&\textbf{99.9}\\ \hline
                              & Ma et al.~\cite{sparse_to_dense}&  0.238&0.055&95.8&99.0&99.7 \\ 
            200                & Nconv-CNN~\cite{cnn_confidence} &  0.180&\textbf{0.030}&97.6&99.4&99.8\\ \cline{2-7}
                              & Ours &  \textbf{0.160}&0.032&\textbf{97.9}&\textbf{99.5}&\textbf{99.9} \\ \hline
   \end{tabular}
   }
   \end{center}
\end{table}

By comparing the results in Table~\ref{tab:nyu_v2_result}, Table~\ref{tab:B3DO} and Table~\ref{tab:sun_3d}, 
we can find that all these three methods yield a little worse results on the dataset collected by Xtion,
which may be caused by different camera intrinsic parameters and the extrinsic parameters between image sensor and depth sensor.
How to design method with better generalization capability between different devices is an interesting direction for the future study.

%%%%%%%%%%%%%
%!TEX root = root.tex

\section{Conclusion}
We propose a guided convolutional network to recover dense depth from sparse and irregular LiDAR points with an RGB image as guidance.
Our novel guided network can dynamically predict content-dependent and spatially-variant kernel weights according to the guidance image to facilitate depth completion.
We further design a convolution factorization to reduce GPU memory consumption
such that our guided convolution module can be applied in powerful encoder-decoder network with multi-stage fusion scheme.
Extensive experiments and ablation studies verify the superior performance of our guided convolutional network and the effectiveness of the feature fusion strategy on depth completion.
Our method not only shows strong results on both indoor and outdoor scenes,
but also presents strong generalization capability under different point densities, various lighting and weather conditions as well as cross-dataset evaluations.
While this paper specifically focuses on the problem of depth completion,
we believe that other tasks in computer vision involving multi-sources as input can also benefit from the design of our guided convolution module and the fusion scheme in our method.

% Can use something like this to put references on a page
% by themselves when using endfloat and the captionsoff option.
\ifCLASSOPTIONcaptionsoff
  \newpage
\fi

% trigger a \newpage just before the given reference
% number - used to balance the columns on the last page
% adjust value as needed - may need to be readjusted if
% the document is modified later
%\IEEEtriggeratref{8}
% The "triggered" command can be changed if desired:
%\IEEEtriggercmd{\enlargethispage{-5in}}

% references section

% can use a bibliography generated by BibTeX as a .bbl file
% BibTeX documentation can be easily obtained at:
% http://mirror.ctan.org/biblio/bibtex/contrib/doc/
% The IEEEtran BibTeX style support page is at:
% http://www.michaelshell.org/tex/ieeetran/bibtex/
%\bibliographystyle{IEEEtran}
% argument is your BibTeX string definitions and bibliography database(s)
%\bibliography{IEEEabrv,../bib/paper}
%
% <OR> manually copy in the resultant .bbl file
% set second argument of \begin to the number of references
% (used to reserve space for the reference number labels box)
% \begin{thebibliography}{1}

\bibliographystyle{IEEEtran}
\bibliography{IEEEabrv,cite}
% \end{thebibliography}
% \bibitem{IEEEhowto:kopka}
% H.~Kopka and P.~W. Daly, \emph{A Guide to \LaTeX}, 3rd~ed.\hskip 1em plus
%   0.5em minus 0.4em\relax Harlow, England: Addison-Wesley, 1999.

% \end{thebibliography}

% biography section
% 
% If you have an EPS/PDF photo (graphicx package needed) extra braces are
% needed around the contents of the optional argument to biography to prevent
% the LaTeX parser from getting confused when it sees the complicated
% \includegraphics command within an optional argument. (You could create
% your own custom macro containing the \includegraphics command to make things
% simpler here.)
%\begin{IEEEbiography}[{\includegraphics[width=1in,height=1.25in,clip,keepaspectratio]{mshell}}]{Michael Shell}
% or if you just want to reserve a space for a photo:

\end{document}